\documentclass{article}
\usepackage{ijcai24}

\usepackage{times}
\usepackage{soul}
\usepackage{url}
\usepackage[hidelinks]{hyperref}
\usepackage[utf8]{inputenc}
\usepackage[small]{caption}
\usepackage{graphicx}
\usepackage{amsmath}
\usepackage{amsthm}
\usepackage{booktabs}
\usepackage{algorithm}
\usepackage{algorithmic}
\usepackage[switch]{lineno}
\usepackage{textcomp}
\usepackage{subfigure}
\usepackage{url}
\usepackage{verbatim}
\usepackage{graphicx}
\usepackage{amssymb}
\usepackage{amsmath}
\usepackage{array,booktabs}
\usepackage{multirow}
\usepackage{bm}
\usepackage{bbding}
\usepackage{pifont}
\usepackage{makecell}
\usepackage{xcolor}
\usepackage{array}
\usepackage{tabularx}
\usepackage{newfloat}
\usepackage{listings}
\usepackage{algorithm}
\usepackage{algorithmic}
\usepackage{hyperref}

\title{A Cognitive-Driven Trajectory Prediction Model for Autonomous Driving in Mixed Autonomy Environments}

\author{
Haicheng Liao\textsuperscript{\rm 1}\thanks{Authors contributed equally; \dag Corresponding author.}\and
Zhenning Li\textsuperscript{\rm 1}$^{*\dag}$\and
Chengyue Wang\textsuperscript{\rm 1}\and
Bonan Wang\textsuperscript{\rm 1}\and
Hanlin Kong\textsuperscript{\rm 2}\and\\
Yanchen Guan\textsuperscript{\rm 1}\and
Guofa Li\textsuperscript{\rm 3}\and
Zhiyong Cui\textsuperscript{\rm 4}\and
Chengzhong Xu\textsuperscript{\rm 1}
\\
\affiliations
$^1$University of Macau\\
$^2$University of Electronic Science and Technology of China\\
$^3$Chongqing University\\
$^4$Beihang University\\
\emails
\{yc27979, zhenningli, chengyuewang, mc3500, yc37976, czxu\}@um.edu.com,
hanlinkong@foxmail.com,
hanshan198@gmail.com,
zhiyongc@uw.edu
}

\begin{document}

\maketitle

\begin{abstract}
As autonomous driving technology progresses, the need for precise trajectory prediction models becomes paramount. This paper introduces an innovative model that infuses cognitive insights into trajectory prediction, focusing on perceived safety and dynamic decision-making. Distinct from traditional approaches, our model excels in analyzing interactions and behavior patterns in mixed autonomy traffic scenarios. It represents a significant leap forward, achieving marked performance improvements on several key datasets. Specifically, it surpasses existing benchmarks with gains of 16.2\% on the Next Generation Simulation (NGSIM), 27.4\% on the Highway Drone (HighD), and 19.8\% on the Macao Connected Autonomous Driving (MoCAD) dataset. Our proposed model shows exceptional proficiency in handling corner cases, essential for real-world applications. Moreover, its robustness is evident in scenarios with missing or limited data, outperforming most of the state-of-the-art baselines. This adaptability and resilience position our model as a viable tool for real-world autonomous driving systems, heralding a new standard in vehicle trajectory prediction for enhanced safety and efficiency.
\end{abstract}

\section{Introduction}
In the evolving landscape of autonomous driving (AD) systems, the challenge of accurately predicting vehicle trajectories is paramount, especially in mixed autonomy environments where autonomous vehicles (AVs) and human-driven vehicles (HVs) dynamically interact. This complex interplay, soon to become an everyday reality on our roads, is marked by its unpredictability, as detailed in studies like \cite{schwarting2019social} and \cite{liao2024bat}. Despite the advancements in deep learning for AD, typified in research such as \cite{sun2023modality} and \cite{geng2023dynamic}, a critical gap remains: these models often lack a nuanced understanding of human driving behavior and decision-making processes \cite{chen2022vehicle,liao2024gpt,guan2024world}. This limitation becomes particularly evident in real-world scenarios where the adaptability of these models to diverse and unforeseen conditions is crucial.

This backdrop prompts us to ask critical questions about the future trajectory of AD: Is the key to advancing AD not just in accumulating more data or refining algorithms, but in gaining a deeper understanding of the driving environment itself? How can we reshape our models to interpret and respond to the intricate human dynamics that underpin driving? Motivated by these questions, our research embarks on an innovative path. We propose a paradigm shift, extending beyond conventional data-driven approaches to embrace a critical yet often-neglected aspect of driving – the concept of \textbf{perceived safety}.
\begin{figure}[t]
  \centering  \includegraphics[width=0.9\linewidth]{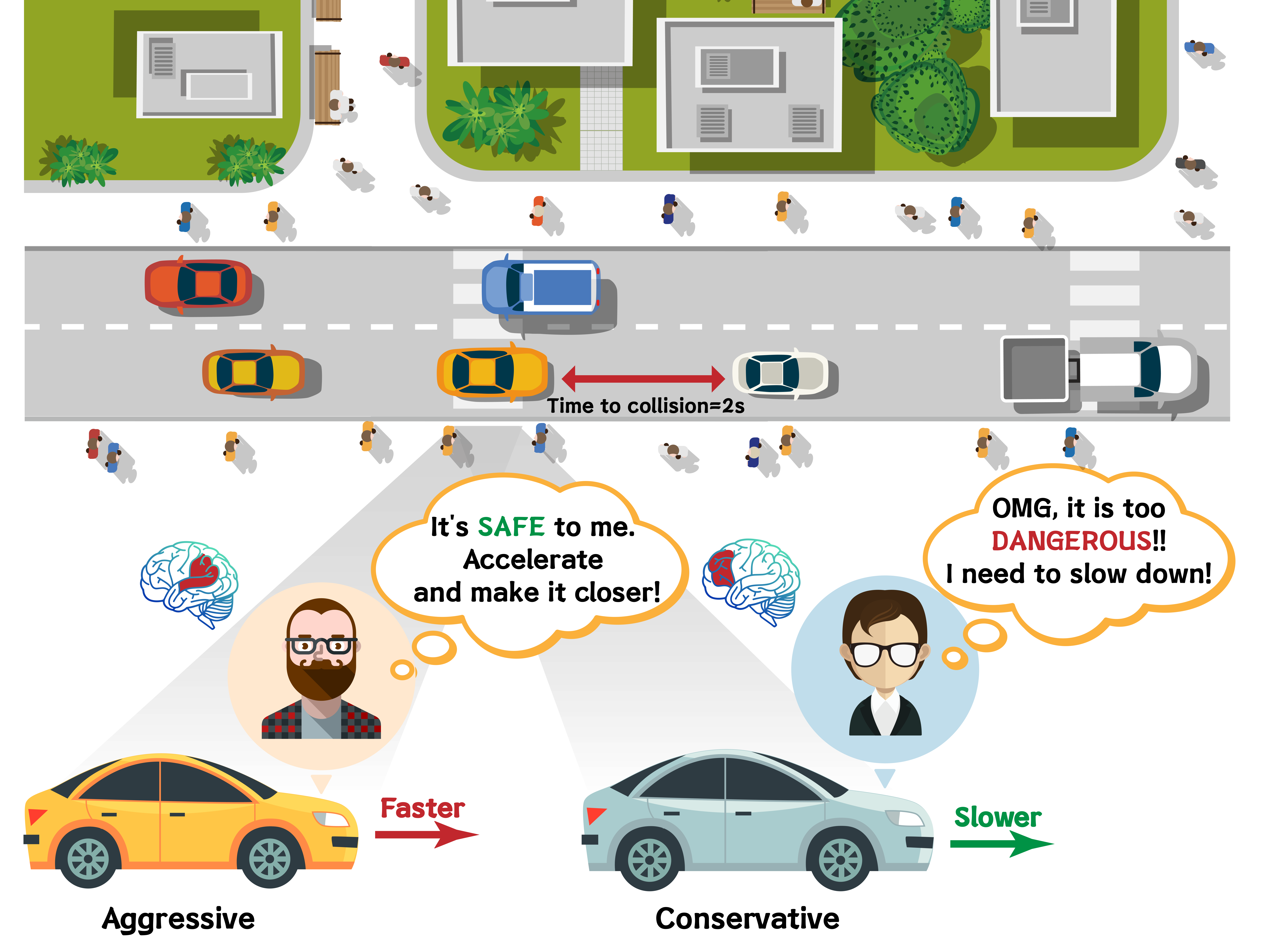} 
  \caption{Perceived safety and its influence on decisions of drivers with different driving behaviors.}
  \label{toutu} 
\end{figure}
This concept, pivotal in shaping driving behaviors and decisions, is deeply rooted in psychological constructs, as detailed in \cite{rubagotti2022perceived}. According to the Theory of Planned Behavior, individual actions in driving are influenced by attitudes (driving behaviors towards others), subjective norms (personal evaluation of safety), and perceived behavioral control (confidence in driving ability) \cite{ajzen1991theory}. Further depth is added by neuroscientific research, such as studies by \cite{saadatnejad2022socially,wei2022fine} and \cite{kronemer2022human}, which unveil that perceived safety is an intricate blend of both conscious and instinctive responses, involving the amygdala's emotional processing and the prefrontal cortex's rational decision-making.
Notably, this nuanced understanding of perceived safety is exemplified in diverse driving scenarios. For instance, when encountering a close car ahead, different drivers exhibit markedly varied responses, as depicted in Figure \ref{toutu}. An aggressive driver, possibly influenced by sensation-seeking tendencies \cite{zuckerman1990psychophysiology}, might quickly swerve, perceiving lower risk. Conversely, a cautious driver, perhaps more risk-averse \cite{rabin2013risk}, might opt for a complete stop. These behaviors, far from being random, are intricately linked to each driver's psychological profile and past experiences, revealing a significant limitation in current AD systems: their inability to account for these complex, cognitive behavioral patterns.

In response, our research integrates the concept of perceived safety into trajectory prediction models for AVs. This integration does more than add a new variable; it injects a human-centric perspective into the heart of these systems. By doing so, we aim to enhance the models' ability to interpret driving behaviors, leading to predictions that are not only technically accurate but also richly informed by contextual and psychological insights. This approach promises a transformative impact on the predictive capabilities of AD systems, aligning them more closely with the multifaceted nature of human driving behavior.

Overall, the key contributions of this study include:

\textbf{1. Perceived Safety-Aware Module Development:}
1) Quantitative Safety Assessment (QSA): A foundational component that objectively evaluates a driving scenario's safety, leveraging metrics like Time-To-Collision (TTC) and Risk Tendency Index (RTI), thereby offering a perception-neutral safety assessment.
2) Driver Behavior Profiling (DBP): Layering upon QSA's foundation, DBP infuses the human element, differentiating between, for instance, an aggressive driver's risk tolerance and a cautious driver's reservations. This profile captures driving nuances in real-time without the need for manual labeling and selection of time windows.

\textbf{2. Social Interaction-Aware Module Development:} 
Acknowledging the collective ballet on roads, we have architected a transformer-based framework, named Leanformer, which encapsulates the intricate inter-vehicular interactions, mirroring contemporary shifts in AD research trajectories.

\textbf{3. Robustness Against Data Inconsistencies:} Our model significantly outperforms the SOTA baseline models when tested on the NGSIM, HighD, and MoCAD datasets, respectively. In a significant stride towards practical applicability, it demonstrates unparalleled resilience in scenarios with incomplete or inconsistent data. This adaptability ensures that our trajectory predictions remain accurate and reliable even when faced with real-world data challenges, a feature often overlooked in conventional models.

\section{Related Work}\label{Related work}
\textbf{Motion Prediction For Autonomous Driving.} A wealth of recent research has harnessed deep learning architectures to predict future motions of target agents for AVs. Employing a variety of frameworks, these models encompass sequential networks \cite{chen2022vehicle,messaoud2021attention}, graph neural networks \cite{rowe2023fjmp}, transformer constructs \cite{chen2022vehicle,yin2021multimodal}, and generative models \cite{LiDCHCG23}. The crux of these approaches is discerning temporal and spatial dynamics between traffic agents from historical data, aiming to optimize prediction accuracy. Moreover, prior research \cite{deo2018convolutional,messaoud2019non,chen2022intention} has tapped into the intricate social interplays among traffic agents, unveiling latent insights and enhancing predictive performance. \\
\textbf{Perceived Safety Concept.} The concept of perceived safety has received extensive attention in psychology and physical human-robot interaction (pHRI) \cite{guiochet2017safety}. In pHRI, perceived safety finds applications in various autonomous physical systems, including mobile robots \cite{scheunemann2020warmth}, industrial manipulators \cite{davis2023role}, humanoid robots \cite{busch2019evaluation}, drones \cite{yao2019autonomous}, and autonomous driving \cite{aledhari2023motion}, to assess and represent people's perception of the level of danger in interactions with robots and their level of comfort during such interactions \cite{bartneck2009measurement}. However, the evaluation of perceived safety in trajectory prediction is often subjective and lacks standard metrics. This work introduces a new quantitative criterion for perceived safety, inspired by human decision-making processes, enhancing the model's contextual understanding of behaviors and traffic conditions.\\
\textbf{Driving Behavior Understanding.}
Numerous previous studies  \cite{schwarting2019social,chandra2020cmetric} have proposed various criteria and metrics to explicitly represent and detect driving behavior, often utilizing classic scales such as Social Value Orientation (SVO) \cite{murphy2011measuring}, Driving Anger Scale (DAS) \cite{deffenbacher1994development}, Driving Anger Expression (DAE), Driving Style Questionnaire (DSQ) \cite{french1993decision}, among others.  In contrast to these traditional methods, our approach models driving behavior in real-time with adaptive, behavior-aware criteria, eliminating the need for manual labeling and offering enhanced learning flexibility. This method addresses challenges related to shifting behaviors and time window selection, reducing data variability associated with discrete classification.

\section{Problem Formulation}\label{Problem Formulation}
Our objective is to predict the trajectories of \textit{target vehicle} (denoted with subscript 0) within the perceptual boundaries of an autonomous vehicle (AV), referred to as the \textit{ego vehicle}, in settings with mixed autonomy. Given the present moment as \( t \), the task is to estimate the probable future trajectory, \( \bm{Y}_{0}^{t: t+t_{f}} \), of the target vehicle over the ensuing \( t_{f} \) time intervals. This estimation is based on the historical states of the target vehicle, \( \bm{X}_{0}^{t-t_{h}:t} \), and the states of surrounding vehicles (indicated by subscript 1:n), \( \bm{X}_{1:n}^{t-t_{h}:t} \), within a defined duration \( t_{h} \). Given the inherent uncertainties in trajectory predictions, our methodology adopts a multimodal prediction framework. This approach discerns various potential maneuvers for the target vehicle and calculates the corresponding probabilities from prior states. As a result, we obtain a spectrum of predictions, each paired with its confidence level.

Central to our model is the use of historical states as inputs. These comprise the 2D position coordinates \( p_{0:n}^{t-t_{h}:t} \), velocity \( v_{0:n}^{t-t_{h}:t} \), and acceleration \( a_{i}^{t-t_{h}:t} \) spanning the duration from \( t-t_{h} \) to \( t \). Formally defined, the input is:
\begin{equation}\label{eq.1}
   \bm{X}_{i}^{t-t_{h}:t} = \left\{p_{i}^{t-t_{h}:t}, v_{i}^{t-t_{h}:t}, a_{i}^{t-t_{h}:t}\right\}, \forall i\in[0,n]
\end{equation}
The resulting output is:
\begin{equation}\label{eq.2}
    \bm{Y}_{0}^{t:t+t_{f}}= \left\{\bm{p}_{0}^{t+1},\bm{p}_{0}^{t+2},\ldots,\bm{p}_{0}^{t+t_{f}-1},\bm{p}_{0}^{t+t_{f}}\right\}
\end{equation}
where $\bm{p}_{0}^{t}=\{({p}_{0,1}^{t}; {c}_{0,1}^{t}),({p}_{0,2}^{t};{c}_{0,2}^{t}),\cdots, ({p}_{0,M}^{t};{c}_{0,M}^{t})\}$ encompasses both the potential trajectory and its associated likelihood ($\Sigma_{1}^{M} {c}_{0,i}^{t}=1$), with $M$ enoting the total number of potential trajectories predicted.

\begin{figure*}[t]
  \centering
\includegraphics[width=0.86\textwidth]{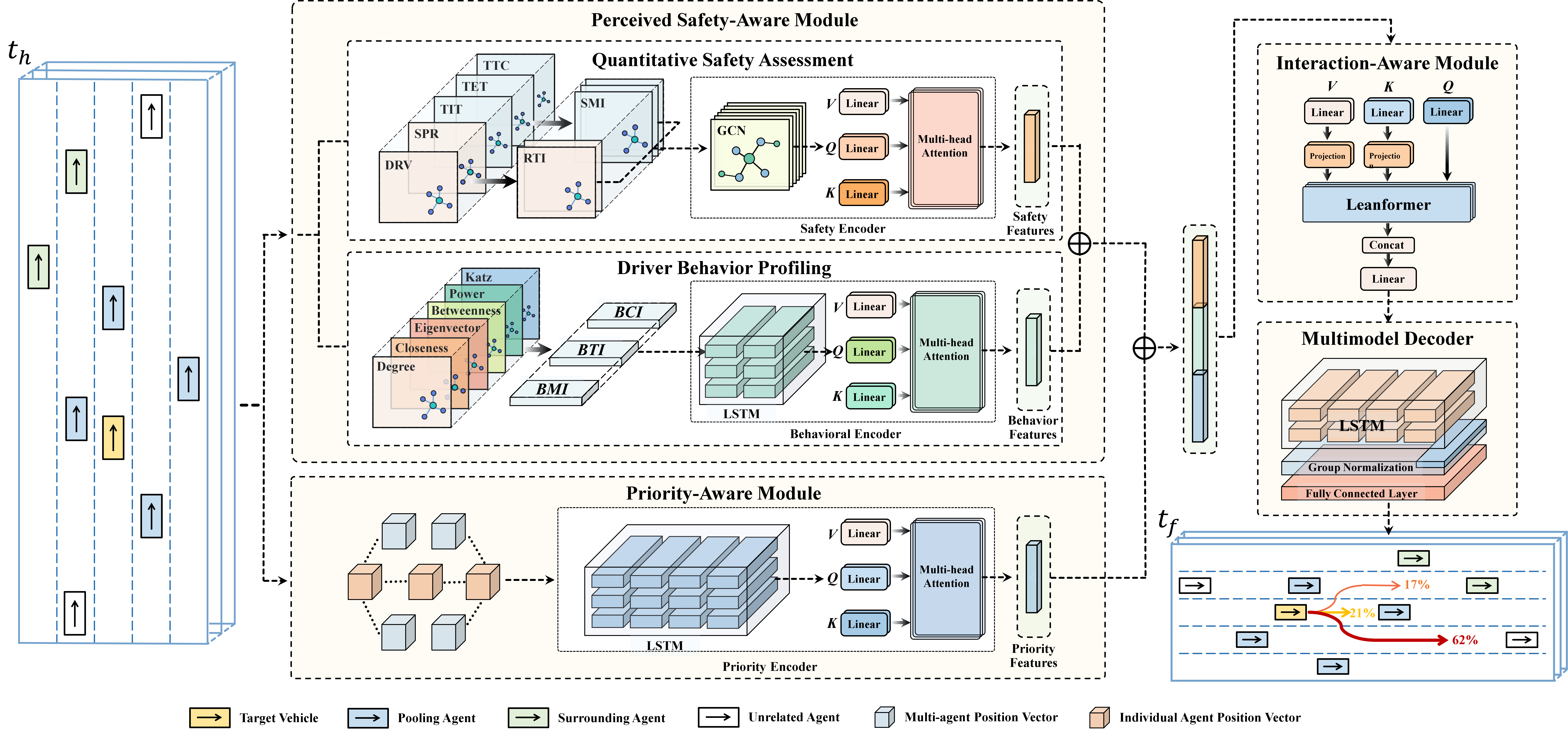} 
  \caption{Architecture of proposed trajectory prediction model.}
  \label{fig1} 
\end{figure*}

\section{Trajectory Prediction Model}\label{Proposed Model}
Figure \ref{fig1} illustrates the architecture of our model. Rooted in the encoder-decoder paradigm, the model seamlessly incorporates three novel modules: the Perceived Safety-Aware, the Priority-Aware, and the Interaction-Aware modules. Collectively, these modules are designed to emulate the human decision-making process during driving.

\subsection{Perceived Safety Aware Module}
\textbf{Quantitative Safety Assessment.} In traffic safety domain, three metrics have gained prominence for their comprehensive portrayal of on-road risks: TTC, Time Exposed Time-to-Collision (TET), and Time Integrated Time-to-Collision (TIT) \cite{minderhoud2001extended}. To align these with our study's objectives, we made slight modifications. Using the 2D position coordinates \( p_{i}^{t} \), \( p_{j}^{t} \) and velocity \( v_{i}^{t} \), \( v_{j}^{t} \) for vehicles \( i \) and \( j \) at time \( t \), these metrics are formulated as:\\
\textit{1) Time-to-Collision (TTC):} TTC is a widely accepted measure used to evaluate the time available before two vehicles collide if they continue on their current trajectories. It offers insights into imminent collision risks and serves as an early warning indicator. TTC for the \( i \)th vehicle is computed as:
\begin{equation}\label{eq.3}
TTC_{i}^{t} = -\frac{d_{i,j}^{t}}{\dot{d}_{i,j}^{t}}
\end{equation}
where \( d_{i,j} \) represents the distance between vehicles \( i \) and \( j \), and \( \dot{d}_{i,j} \) is its rate of change:
\begin{equation}\label{eq.4}
    \left\{\begin{array}{l}
d_{i,j}^{t} =\sqrt{\left({p}_i^{t}-{p}_j^{t}\right)^{\top}\left({p}_i^{t}-{p}_j^{t}\right)} \\
\dot{d}_{i,j}^{t} =\frac{1}{d_{i,j}^{t}}\left({p}_i^{t}-{p}_j^{t}\right)^{\top}\left({v}_i^{t}-{v}_j^{t}\right)
\end{array}\right.
\end{equation}\\
\textit{2) Time Exposed Time-to-Collision (TET):} It measures the exposure duration to critical TTC values within \( t_h \). It is the sum product of a switching variable and a time threshold $\tau_{\mathrm{sc}}$ (set at 0.1s):
\begin{equation}\label{eq.9}
TET_i^{t_{k}}=\sum_{{t_{k}}=t-t_{h}}^{t} \delta_i(t_{k}) \cdot \tau_{\mathrm{sc}}
\end{equation}
with the switching variable given by:
\begin{equation}
\delta_i(t_{k})=\left\{\begin{array}{lc}
1 & \forall \ \ \ 0  \leq \mathrm{TTC}_i^{t_{k}} \leq \mathrm{TTC}^*\\
0 & \text { otherwise }
\end{array}\right.
\end{equation}
In our study, \( \mathrm{TTC}^* \), delineating safe and critical thresholds, is set at 3.0s.\\
\textit{3) Time Integrated Time-to-Collision (TIT):}
An adaptation of TET, TIT integrates TTC profile to evaluate safety levels, factoring in the evolution of each vehicle's TET temporally:
\begin{equation}
TIT_i^{{t}_{k}}=\sum_{\tilde{t}=t-t_{f}}^{t_{h}}\left[\mathrm{TTC}^*-\mathrm{TTC}_{i}({t}_{k})\right] \cdot \tau_{\mathrm{sc}}
\end{equation}
Elevated values of TTC, TET, and TIT imply sustained exposure to potential collision risks, underscoring a deterioration in perceived safety.\\
\textit{4) Risk Tendency Index (RTI):} To further capture congestion patterns in complex traffic environments, we propose an index between the $i$th and $j$th vehicles at time $t$, denoted as \textit{subjective risk perception indicator} (SPR), i.e. ${ {R}}_{i}^{{t}}$, and \textit{dynamic risk volatility indicator} (DRV), i.e. $ {R}^{{t}}_{i,j}$, respectively:
\begin{equation}\label{eq.12}
{ {R}}_{i}^{{t}}= {R}_{j}^{{t}}=\left[ {R}^{{t}}_{i,j},  {\dot{R}}^{{t}}_{i,j}\right]^{T}, \forall i, j \in[0,n], i \neq j 
\end{equation}
In this context, the vector ${R}^{{t}}_{i,j}$ with larger values indicates an increased risk of collision, while the vector $\dot{ {R}^{{t}}}_{i,j}$ characterizes the dynamic congestion conditions in complex traffic scenarios. In addition, the definitions of SPR and DRV are defined as follows:
\begin{equation}
 {R}^{{t}}_{i,j}= {R}^{{t}}_{j,i}=\left\{\begin{array}{l}
1 / e^{{q}_{i,j}^{{t}}}, {q}_{i,j}^{{t}}>0 \\
0, {q}_{i,j}^{{t}}=0
\end{array} \right. 
\end{equation}
where the DRV $ {\dot{R}}^{{t}}_{i,j}$ represents the gradient to evaluate fluctuations in SPR $ {{R}}^{{t}}_{i,j}$ and can be expressed as follows:
\begin{equation}
\dot{ {R}^{{t}}}_{i,j}=\dot{ {R}}^{{t}}_{j,i}=\left\{\begin{array}{l}
1 / e^{\dot{{q}}_{i,j}^{{t}}}, \dot{{q}}_{i,j}^{{t}}>0 \\
0, \dot{{q}}_{i,j}^{{t}}=0
\end{array}\right. 
\end{equation}
The quantities ${q}_{i,j}^{{t}}$ and $\dot{q}_{i,j}^{{t}}$ are calculated based on several critical parameters related to the dynamics of two traffic agents. These parameters include the lateral velocity $v_{x}^{t}$, longitudinal velocity $v_{y}^{t}$, 2D position coordinate $ p_{x}^{t}$ and $ p_{y}^{t}$,  as well as the lateral  $a_{x}^{t}$and longitudinal $a_{y}^{t}$. Mathematically, it can be represented as follows:
\begin{equation}
   {q}_{i,j}^{{t}} = \max \left(-\frac{\Delta_{i,j} v_{x}^{t} \times \Delta_{i,j} p_{x}^{t} +\Delta_{i,j} v_{y}^{t} \times \Delta_{i,j} p_{y}^{t}}{\Delta_{i,j} v_{x}^2+\Delta_{i,j} v_{y}^2}, 0\right)
\end{equation}
\begin{equation}
     \dot{q}_{i,j}^{{t}} = -\frac{\Delta_{i,j} a_{x}^{t} \times \Delta_{i,j} p_{x}^{t} +\Delta_{i,j} a_{y}^{t} \times \Delta_{i,j} p_{y}^{t}}{\Delta_{i,j} a_{x}^2+\Delta_{i,j} a_{y}^2}
\end{equation}
where the $\Delta_{i,j} (\cdot)$ denotes the difference between quantities of the $i$-th and $j$-th vehicles.
A larger vector $ {R}^{{t}}_{i,j}$ indicates a higher risk of collision, while the vector $\dot{ {R}^{{t}}}_{i,j}$ describes the dynamic congestion conditions in complex traffic scene.

The encoder within the Perceived Safety-Aware module combines the GCNs \cite{hamilton2017inductive} and the scaled dot-product multi-head self-attention mechanism \cite{vaswani2017attention}.
To capture the dynamic geometric relationships among traffic agents, we employ a convolutional neural network on a fully connected interaction multigraph. This multigraph operation layer incorporates sequential perceived-safety criteria $ {\bm{Q}}$ as nodes and adjacency matrix $A$  as graph edges. The GCN layer is given by:
\begin{equation}
\bm{Z}^{k+1}_i=\phi_{\text {ReLU}}\left(\tilde{\bm{D}}^{-\frac{1}{2}} \tilde{\bm{A}} \tilde{\bm{D}}^{-\frac{1}{2}} \bm{Z}^k_i \bm{W}^k_i\right)
\end{equation}
where $\tilde{\bm{D}}$ is the degree matrix.
The multi-head self-attention mechanism then transform the feature matrix $\bm{Z}^{k+1}_i$ output from GCNs into query, key, and value vectors. The output for the \(i\)th agent from this mechanism is denoted as \( {\alpha}\).

For improved training stability, inspired by ResNet \cite{he2016deep}, we integrate Gate Linear Units (GLUs) \cite{dauphin2017language} and Layer Normalization \cite{ba2016layer} to efficiently manage features:
\begin{flalign}
 {\bar{O}}^{t-t_{h}:t} &=  { \phi_{\textit {LN}}}\left(  { \phi_{\textit {MLP}}}( { \phi_{\textit {GLUs}}}({\alpha}))\right)
\end{flalign}
In particular, the GLUs layer can be defined as:
\begin{equation}
    \phi_{\textit{GLUs}}(\alpha) = ( \alpha W_1 + b_1 ) \odot \phi_{\textit{sigmoid}}( \alpha W_2 + b_2 )
\end{equation}
where \( \alpha \) represents the safe attention coefficient from the multi-head attention mechanism, \( W_1 \) and \( W_2 \) are the learnable weight parameters associated with the GLUs layer, \( b_1 \) and \( b_2 \) are the corresponding biases, \( \odot \) denotes element-wise multiplication, \( \phi_{\textit{sigmoid}} \) is the sigmoid activation function, and \( \phi_{\textit{LN}}(\cdot) \) stands for Layer Normalization. Correspondingly, the output of the encoder within the QSA is denoted as ${\bar{O}}^{t-t_{h}:t}_{safe}$.\\
\textbf{Driver Behavior Profiling.} \label{Behavior-aware Module_0}
Unlike traditional methods that classify driver behavior into fixed, predefined categories, we propose a novel dynamic geometric graph (DGG) based real-time profiling approach. This continuum solution addresses the problem of fluctuating behavioral categorization in previous studies. At a given time \( t \), the graph \( G^{t} \) is constructed as
${G}^{t} = \{V^{t},{E}^{t}\}$, where \( V^{t}=\{{v}_{0}^{t},{v}_{1}^{t},...,{v}_{n}^t\} \) signifies the node set, and each node \( {v}_{i}^{t} \) represents an individual agent.
\( {E^{t}} = \{{e_{0}^{t}},{e_{1}^{t}},...,{e_{n}^{t}}\} \) is the set of edges illustrating potential interactions between agents. Specifically, an edge \( {e_{i}^{t}} \) connects node \( {v}_{i}^{t} \) to other nodes (agents) potentially influencing it.

Interactions are considered when agents, such as \( v_{i} \) and \( v_{j} \), are sufficiently close. Specifically, if their distance \( d(v_{i}^{t}, v_{j}^{t}) \) is within a set threshold \( r \), an interaction is inferred. This relationship is represented as ${e_{i}^{t}}= \{{v_{i}^{t}}, {v_{j}^{t}} \mid(j \in  {N}_{i}^{t})\}$ where the neighborhood set \( {N}_{i}^{t} \) is defined by:
\begin{equation}
    {N}_{i}^{t}=\{v_{j}^{t} \in V^{t}\setminus\{v_{i}^{t}\} \mid d( v_{i}^{t}, v_{j}^{t}) \leq r, \text{ and } i \neq j\} 
\end{equation}
The adjacency matrix \( A^{t} \) of \( G^{t} \) is symmetric. Formally,
\begin{equation}\label{eq.8}
    A^{t}(i, j)= \begin{cases}
    d(v_{i}^{t}, v_{j}^{t}) & \text{if } d(v_{i}^{t}, v_{j}^{t})\leq{r} \text{ and } i \neq j \\
    0 & \text{otherwise}
    \end{cases}
\end{equation}

To further evaluate the behavior of individual agents and reveal important agents and overall connectivity in the traffic graph, centrality measures are used, including degree $ {J}_{i}^{t}(D)$, closeness $ {J}_{i}^{t}(C)$, eigenvector ${J}_{i}^{t}(E)$, betweenness ${J}_{i}^{t}(B)$, power ${J}_{i}^{t}(P)$, and Katz ${J}_{i}^{t}(K)$ centrality. Formally,\\
\textit{(1) Degree Centrality:} Reflects an agent's number of connections, showing its influence on and vulnerability to others.
\begin{equation}
    {J}_{i}^{t}(D)= \left|  {N}_{i}^{t}\right|+ {J}_{i}^{t-1}(D)
\end{equation}\\
\textit{(2) Closeness Centrality:} Indicates an agent's reachability, suggesting its potential influence over others:
\begin{equation}
    {J}_{i}^{t}(C)=\frac{\left|  {N}_{i}^{t}\right|-1}{\sum_{\forall v_{j}^{t} \in  {N}^{t}_{i}}d\left(v_{i}^{t}, v_{j}^{t}\right)}
\end{equation}\\
\textit{(3) Eigenvector Centrality:} Measures an agent's importance by considering both quantity and quality of connections:
\begin{equation}
    {J}_{i}^{t}(E)=\frac{ \sum_{\forall v_{j}^{t} \in  {N}^{t}_{i}}d\left(v_{i}^{t}, v_{j}^{t}\right)}{\lambda}
\end{equation}\\
\textit{(4) Betweenness Centrality:} Highlights agents that act as bridges or bottlenecks in traffic, crucial in congested scenes:
\begin{equation}
    {J}_{i}^{t}(B) = \sum_{\forall v_{s}^{t},v_{k}^{t} \in {V}^{t}} \frac{\sigma_{j,k}(v_{i}^{t})}{\sigma_{j,k}}
\end{equation}\\
\textit{(5) Power Centrality:} Identifies agents in recurrent interactions, hinting at traffic patterns:
\begin{equation}
    {J}_{i}^{t}(P) = \sum_{k}\frac{A^{k}_{ii}}{k!}
\end{equation}\\
\textit{(6) Katz Centrality:} Emphasizes both direct and distant interactions of an agent, capturing intricate driving patterns:
\begin{scriptsize}
\begin{equation}
    {J}_{i}^{t}(K) = \sum_{k} \sum_{j} \alpha^{k} A^{k}_{ij}+\beta^{k},  \forall i, j \in[0,n], \text { where } \alpha^{k} <\frac{1}{\lambda_{\max }}
\end{equation}
\end{scriptsize}

To provide better performance in profiling driving behavior, we draw inspiration from metrics such as speed, acceleration, and jerk. Based on these, we introduce three continuous criteria: Behavior Magnitude Index (BMI) ${\mathcal{C}}_{i}^{t}$, which measures the influence of driving behaviors by evaluating their centrality; Behavior Tendency Index (BTI) ${\mathcal{L}}_{i}^{t}$, which quantifies behavior propensity by calculating time series derivatives, suggesting higher probabilities of specific behaviors with larger derivatives; and Behavior Curvature Index (BCI) ${{\mathcal{I}}}_{i}^{t}$, which uses the jerk concept to measure the intensity of driving behaviors by calculating the second-order derivatives of continuous centrality measures. Formally,
\begin{equation}
{\mathcal{C}}_{i}^{t}=\left|{ \mathcal{J}^{t}_{i}(\tilde{X}_i^t)}\right|^{T},{\mathcal{L}}_{i}^{t}=\left|\frac{\partial{{{\mathcal{C}}}}_{i}^{t}}{\partial t}\right|^{T}, {{\mathcal{I}}}_{i}^{t} = \left|\frac{\partial {\mathcal{C}}_{i}^{t}}{\partial^{2} t}\right|^{T}
\end{equation}
where $\tilde{X}_i^t =\left[ {J}_{i}^{t}(D), {J}_{i}^{t}(C), {J}_{i}^{t}(E), {J}_{i}^{t}(B), {J}_{i}^{t}(P), {J}_{i}^{t}(K)\right]$.

In combination, these criteria offer a holistic view of driving behaviors without needing manual labeling during training, addressing issues of changing behavior labels and time window selection. 

The encoder of Driver Behavior Profiling comprises two main components: the LSTM and multi-head self-attention mechanism. The behavior-aware criteria are first processed by the LSTM, yielding temporal vectors:
\begin{equation}
 {\tilde{J}}_{i}^{t-t_{h}:t}=\phi_{ \textit{LSTM}}\left({{h}}^{t-t_{h}:t}_{i},\phi_{\textit{MLP}}( {J}_{i}^{t-t_{h}:t}), \phi_{\textit{MLP}}( {\bar{O}_{\textit{safety}}}^{t-t{h}:t})\right)
\end{equation}
Here, the LSTM encoder updates the hidden state of agent \( v_i \) frame-by-frame with shared weights. Using the multi-head self-attention mechanism and GLUs, we compute attention weights for different agent behaviors, producing precise sequential behavioral features in a similar way in QSA:
\begin{equation}
 {\bar{O}_{\textit{behavior}}}^{t-t_{h}:t} = \phi_{\textit {LN}}\left( \phi_{\textit {MLP}}(\phi_{\textit {GLUs}}({\alpha}^{\textit{behavior}}))\right)
\end{equation}

Lastly, the behavioral features are integrated with priority features and contextual features, then forwarded to the Interaction-Aware module for further processing.

\subsection{Priority-Aware Module}\label{Priority-Aware Pooling Module_0}
In light of recent advances in cognitive studies, it has become evident that the spatial positioning of vehicles within a scene can variably influence the behavior and decisions of a target vehicle. For instance, vehicles located directly in the anticipated trajectory path tend to exert greater influence relative to those situated behind. Furthermore, during overtaking maneuvers, vehicles positioned on the left may carry augmented significance. Recognizing these spatial intricacies, we introduce the Priority-Aware Module. This sophisticated module adeptly transforms the spatial coordinates of agents, encoding them into high-dimensional positional vectors, thereby yielding detailed positional features.

Our pooling mechanism adeptly amalgamates dynamic positional information from the encompassing traffic scenario, effectively capturing both singular and multi-agent positional vectors. This mechanism emphasizes the dynamic nuances of position data, accommodating historical agent states, denoted as \( \bm{S}_{i}^{t_{k}} \), as well as the intricate spatial interplay symbolized by \( \bm{P}_{i, j}^{t_{k}} \). Mathematically, these relationships are represented as
$\bm{S}_{i}^{t_{k}} = \{ p_{i}^{t_{k}} - p_{i}^{t_{k}-1}, v_{i}^{t_{k}} - v_{i}^{t_{k}-1}, a_{i}^{t_{k}} - a_{i}^{t_{k}-1} \}$ and
$\bm{P}_{i, j}^{t_{k}} = \{ p_{i}^{t_{k}} - p_{j}^{t_{k}}, v_{i}^{t_{k}} - v_{j}^{t_{k}}, a_{i}^{t_{k}} - a_{j}^{t_{k}} \}$, where $t_{k} \in [t-t_{h},t]$. Within this module, the encoder, which amalgamates both LSTM and multi-head attention mechanisms, meticulously processes the dynamic positional vectors. This processing involves transmuting discrete position vectors into a more continuous spatio-temporal domain, thereby enhancing the representation of temporal and spatial dynamics. At each discrete temporal instance \( t \), the encoder assimilates recent historical position vectors via an LSTM network:
\begin{equation}
 {{O}}_{\textit{priority}}^{t-t_{h}:t}=\phi_{ \textit{LSTM}}\left(\bm{\bar{h}}^{t-t_{h}:t-1}_{i}, \bm{S}^{t-t_{h}:t-1}_{i}, \bm{{P}}^{t-t_{h}:t-1}_{i,j}\right)
\end{equation}

Subsequently, the LSTM's output is channeled through a multi-head attention mechanism, culminating in the synthesis of refined priority features:
\begin{equation}
 {\bar{O}_{\textit{priority}}}^{t-t_{h}:t} =  { \phi_{\textit {LN}}}\left(  { \phi_{\textit {MLP}}}( { \phi_{\textit {GLUs}}}({\alpha}^{\textit{ priority}}))\right)
\end{equation}

\subsection{Interaction-Aware Module}\label{Interaction-Aware Pooling Module}
To better understand the synergistic influence of surrounding vehicles' risk levels, positions, and one's own behavior on the target vehicle's future trajectory, we introduce the Leanformer in this module, which is a lightweight version of the Transformer, allowing for a favorable trade-off between accuracy and efficiency. Mathematically, 
\begin{equation}
     {{O}}^{t-t_{h}:t} = \phi_{\textit {Leanformer}}\left( {\bar{O}_{\textit{safety}}}^{t-t_{h}:t}\| {\bar{O}_{\textit{behavior}}}^{t-t_{h}:t}\| {\bar{O}_{\textit{ priority}}}^{t-t_{h}:t}\right)
\end{equation}
This equation captures the integrated effect of safety perception, behavioral tendencies, and spatial positioning on the target vehicle's trajectory.

\subsection{Multimodal Decoder}
The decoder, rooted in a Gaussian Mixture Model (GMM) with multimodality, employs a dedicated LSTM and a fully connected layer. It processes the composite interactive vectors \( {\bar{O}} \) to forecast the target vehicle's trajectory. The predicted trajectory, \( \boldsymbol{\bm{Y}_{0}^{t:t+t_{f}}} \), is determined by:
\begin{equation}
    \boldsymbol{\bm{Y}_{0}^{t:t+t_{f}}} = {F}_{\theta}\left( {F}_{\theta}( {\bar{O}})\right)
\end{equation}
where ${F}_{\theta}(\bm{\cdot})=\phi_{\textit{ReLU}}\left(\phi_\textit{MLP}\left[\phi_{\text {GN}}\left(\phi_{\textit {LSTM}}(\bm{\cdot})\right)\right]\right)$
Moreover, \( \phi_{\text {ReLU}} \) is the ReLU activation, and \( \phi_{\text{GN}} \) is Group Normalization, used for improved training stability. The decoder's output comprises multiple future trajectories for target vehicle.

\section{Experiments}\label{Experiments}

\subsection{Experimental Setup}
We carried out experiments using three prominent datasets: NGSIM \cite{deo2018convolutional}, HighD \cite{8569552}, and MoCAD \cite{liao2024bat}, adhering to a common dataset division framework. Trajectories were split into 8s segments, where the initial 3s served as historical data, and the succeeding 5s were used for assessment, forming the \textit{complete} test set. The Root Mean Square Error (RMSE) was employed as the evaluation metric. Recognizing a gap in research concerning data omissions in trajectory prediction,  we established the \textit{missing} test set, which was further categorized into three subsets based on the duration of data omissions: \textit{drop 3-frames}, \textit{drop 5-frames}, and \textit{drop 8-frames}. Omissions were purposefully made around the midpoint of the historical trajectory. For instance, in the \textit{drop 5-frames} subset, data ranging from the $(t-8)$th to the $(t-12)$th frame was excluded. To manage these omissions, we employed simple linear interpolation. Furthermore, we trained our model on only 25\% of the dataset and evaluated it on \textit{complete} test sets to evaluate its ability to handle unfamiliar data and manage data omissions. Additionally, our model is trained until convergence on four A40 48G GPUs. We employ the Adam optimizer and utilize CosineAnnealingWarmRestarts for the scheduler. The model is trained with a batch size of 64, and the learning rates are set from $10^{-3}$ to $10^{-5}$. Building on the insights of Kendall et al. \cite{kendall2018multi}, we incorporated a multitask learning paradigm into our model's loss function.  We fuse the RMSE and the Negative Log-Likelihood criterion to constitute the comprehensive loss function.
\begin{table}[htbp]
  \centering
  \caption{Evaluation of the proposed model and baselines on the NGSIM dataset's \textit{complete} and \textit{missing} test sets over a 5-second prediction horizon. The accuracy metric is RMSE (m). Cases marked as ('-') indicate unspecified values. \textbf{Bold} and \underline{underlined} values represent the best and second-best performance in each category.}
  \resizebox{0.95\linewidth}{!}{
  \setlength{\tabcolsep}{2mm}
    \begin{tabular}{cccccc}
    \toprule
    \multirow{2}[3]{*}{Model} & \multicolumn{5}{c}{Prediction Horizon (s)} \\
\cmidrule{2-6}          & 1     & 2     & 3     & 4     & 5 \\
    \midrule
    S-LSTM \cite{alahi2016social}& 0.65  & 1.31  & 2.16  & 3.25  & 4.55  \\
    S-GAN \cite{gupta2018social}& 0.57  & 1.32  & 2.22  & 3.26  & 4.40  \\
    CS-LSTM \cite{deo2018convolutional}& 0.61  & 1.27  & 2.09  & 3.10  & 4.37  \\
    MATF-GAN \cite{zhao2019multi}& 0.66  & 1.34  & 2.08  & 2.97  & 4.13  \\
    DRBP\cite{gao2023dual}& 1.18  & 2.83  & 4.22  & 5.82  & - \\
    MFP \cite{tang2019multiple}& 0.54  & 1.16  & 1.89  & 2.75  & 3.78  \\
    NLS-LSTM \cite{messaoud2019non}& 0.56  & 1.22  & 2.02  & 3.03  & 4.30  \\
    MHA-LSTM \cite{messaoud2021attention}& 0.41  & 1.01  & 1.74  & 2.67  & 3.83  \\
    WSiP \cite{wang2023wsip}& 0.56  & 1.23  & 2.05  & 3.08  & 4.34  \\
    CF-LSTM \cite{xie2021congestion}& 0.55  & 1.10  & 1.78  & 2.73  & 3.82  \\
    TS-GAN \cite{wang2022multi}& 0.60  & 1.24  & 1.95  & 2.78  & 3.72  \\
    STDAN \cite{chen2022intention}& 0.42  & 1.01  & 1.69  & 2.56  & 3.67  \\
    iNATran \cite{chen2022vehicle} & 0.39  &0.96  &1.61  & 2.42  & 3.43  \\
     BAT (25\%) \cite{liao2024bat}& \underline{0.31}  & \underline{0.85}  &1.65  & 2.69  & 3.87 \\
    BAT \cite{liao2024bat}& \textbf{0.23} & \textbf{0.81}  & {1.54}  & 2.52 &3.62\\
    FHIF \cite{zuo2023trajectory} &0.40  & 0.98  & 1.66  & 2.52  & 3.63\\ 
    DACR-AMTP \cite{cong2023dacr}& 0.57  & 1.07  & 1.68  & 2.53  & 3.40 \\ 
    \midrule
    \textbf{Our Model} & \underline{0.31} & \textbf{ 0.81 } & \textbf{ 1.44 } & \textbf{ 2.08 } & \textbf{ 2.85 } \\
    Our Model (25\%) & 0.43  & 0.94  & 1.57  & 2.55  & 3.32  \\
    Our Model (drop 3-frames) & 0.40  & 0.88  & \underline{1.53}  & \underline{2.34}  & \underline{2.99}  \\
    Our Model (drop 5-frames) & 0.44  & 0.92  & 1.57  & 2.42  & 3.34  \\
    Our Model (drop 8-frames) & 0.48  & 0.98  & 1.68  & 2.47  & 3.52  \\
    \bottomrule
    \end{tabular}%
    }
  \label{table_3}%
\end{table}%

\subsection{Experimental Results}
\subsubsection{Performance Comparison on \textit{Complete} Test Set}
As illustrated in Table \ref{table_3}, our model surpasses the best of SOTA baselines with noteworthy gains of 10.6\% and 16.2\% for short-term (1s-3s) and long-term predictions (4s-5s) respectively on the NGSIM dataset. Moreover, Table \ref{table_4} highlights fewer prediction inaccuracies on the HighD dataset than NGSIM, due to HighD's superior trajectory precision and larger sample size. Notably, our model's long-term forecasting outperforms SOTA baselines, showing RMSE improvements of 27.4\% for a 5s prediction horizon. Additionally, as shown in Table \ref{table_6}, our model notably excels on busy urban roads, surpassing SOTA baselines by at least 12.2\% for short-term predictions and reducing long-term prediction errors by at least 0.57 meters on MoCAD, potentially decreasing traffic accident risks. 

\begin{table}[t]
  \centering
  \caption{Evaluation of our model and SOTA baselines on HighD.}
  \setlength{\tabcolsep}{2.5mm}
   \resizebox{0.95\linewidth}{!}{
    \begin{tabular}{cccccc}
    \toprule
    \multirow{2}[3]{*}{Model} & \multicolumn{5}{c}{Prediction Horizon (s)} \\
\cmidrule{2-6}          & 1     & 2     & 3     & 4     & 5 \\
    \midrule
    S-LSTM \cite{alahi2016social}& 0.22  & 0.62  & 1.27  & 2.15  & 3.41  \\
    S-GAN \cite{gupta2018social}& 0.30  & 0.78  & 1.46  & 2.34  & 3.41  \\
    WSiP \cite{wang2023wsip}& 0.20  & 0.60  & 1.21  & 2.07  & 3.14  \\
    CS-LSTM \cite{deo2018convolutional}& 0.22  & 0.61  & 1.24  & 2.10  & 3.27  \\
    MHA-LSTM \cite{messaoud2021attention}& 0.19  & 0.55  & 1.10  & 1.84  & 2.78  \\
    NLS-LSTM \cite{messaoud2019non}& 0.20  & 0.57  & 1.14  & 1.90  & 2.91  \\
    DRBP\cite{gao2023dual}& 0.41  & 0.79  & 1.11  & 1.40  & - \\
    EA-Net \cite{cai2021environment} & 0.15  & 0.26  & 0.43  & 0.78  & 1.32  \\
    CF-LSTM \cite{xie2021congestion}& 0.18  & 0.42  & 1.07  & 1.72  & 2.44  \\
    STDAN \cite{chen2022intention}& 0.19  & 0.27  & 0.48  & 0.91  & 1.66  \\
    iNATran \cite{chen2022vehicle}& 0.04  & \textbf{0.05}  & 0.21  & 0.54  & 1.10 \\
    DACR-AMTP \cite{cong2023dacr} & 0.10  & 0.17  & 0.31  & 0.54  & 1.01  \\
    BAT (25\%) \cite{liao2024bat} & 0.14  & 0.34  &0.65  & 0.89  & 1.27 \\
    BAT \cite{liao2024bat} & {0.08}  & 0.14  & \underline{0.20}  & {0.44} &{0.62}\\
      GaVa \cite{liao2024human}& 0.17  & 0.24  & 0.42  & 0.86  & 1.31  \\ 
    \midrule
    \textbf{Our Model} & \textbf{ 0.04 } & \underline{0.09} & \textbf{ 0.19 } & \textbf{ 0.31 } & \textbf{ 0.45 } \\
    Our Model (25\%) & 0.09  & 0.22  & 0.41  & 0.64  & 0.95  \\
    Our Model (drop 3-frames) & \underline{0.05}  & 0.12  & 0.21  & \underline{0.33}  & \underline{0.49}  \\
    Our Model (drop 5-frames) & 0.17  & 0.31  & 0.44  & 0.65  & 0.98  \\
    Our Model (drop 8-frames) & 0.18  & 0.47  & 0.84  & 1.27  & 1.74  \\
    \bottomrule
    \end{tabular}%
    }
  \label{table_4}%
\end{table}%

\subsubsection{Performance Comparison on \textit{Missing} Test Set}
Tables \ref{table_3}, \ref{table_4}, and \ref{table_6} spotlight our model's robustness, even in the face of incomplete datasets. Consistently, across the \textit{drop 3-frames} and \textit{drop 5-frames} datasets, it outperforms all baselines. Notably, within the \textit{drop 3-frames} context, our model even eclipses several leading SOTA benchmarks that were evaluated using \textit{complete} test sets. However, no model is without its exceptions. In the \textit{drop 5-frames} dataset, while our model largely dominates, it trails behind DACR-AMTP on NGSIM and the short-term forecasts of iNATran on HighD. Still, these are isolated instances, and the overarching trend confirms the model's robustness and versatility. A deeper dive reveals a predictable trend: the model's performance is inversely proportional to the omission of input data. Yet, even under a stringent scenario like the \textit{drop 8-frames}—where half the data is missing—our model stands its ground, frequently matching or surpassing most SOTA benchmarks. This resilience not only underscores its robust forecasting capabilities but also positions it as an invaluable tool for AV training, especially in real-world scenarios riddled with data inconsistencies.

\subsubsection{Performance on a Limited 25\% Training Set}
To challenge our model's adaptability, we trained it using only a quarter of the available training set from the NGSIM, HighD, and MoCAD datasets, yet evaluated its performance on the \textit{complete} test set. Impressively, as shown in Tables \ref{table_3}, \ref{table_4}, and \ref{table_6}, even with this limited training data, our model delivered RMSE values that were notably lower than most baseline models. Such results underscore our model's efficiency and robustness in trajectory prediction. This performance indicates a promising potential: our model might substantially cut down on the data demands typically associated with training autonomous vehicles, particularly in scenarios that are data-scarce or unconventional. In summary, our findings attest to the model's reliability, resource efficiency, and precision in prediction.  

\begin{table}[t]
  \centering
  \caption{Evaluation of our model and SOTA baselines on MoCAD.}
  \setlength{\tabcolsep}{3mm}
  \resizebox{\linewidth}{!}{
    \begin{tabular}{cccccc}
    \toprule
    \multirow{2}[3]{*}{Model} & \multicolumn{5}{c}{Prediction Horizon (s)} \\
\cmidrule{2-6}          & 1     & 2     & 3     & 4     & 5 \\
    \midrule
    S-LSTM \cite{alahi2016social} & 1.73  & 2.46  & 3.39  & 4.01  & 4.93 \\
    S-GAN \cite{gupta2018social} & 1.69  & 2.25  & 3.30  & 3.89  & 4.69  \\
    CS-LSTM \cite{deo2018convolutional} & 1.45  & 1.98  & 2.94  & 3.56  & 4.49  \\
    MHA-LSTM \cite{messaoud2021attention} & 1.25  & 1.48  & 2.57  & 3.22  & 4.20  \\
    NLS-LSTM \cite{messaoud2019non} & 0.96  & 1.27  & 2.08  & 2.86  & 3.93\\
    WSiP \cite{wang2023wsip} & 0.70  & 0.87  & 1.70  & 2.56  & 3.47  \\
    CF-LSTM \cite{xie2021congestion} & 0.72  & 0.91  & 1.73  & 2.59  & 3.44 \\
    STDAN \cite{chen2022intention} & 0.62  & 0.85  & 1.62  & 2.51  & 3.32  \\
    BAT (25\%) \cite{liao2024bat} & 0.65  & 0.99  & 1.89  & 2.81  & 3.58 \\
    BAT \cite{liao2024bat} & \underline{0.35}  & \underline{0.74}  & {1.39}  & {2.19} &{2.88}\\
    HLTP \cite{10468619}  & {0.55} & {0.76} & {1.44} & {2.39} & {3.21} \\
    \midrule
   \textbf{Our Model} & \textbf{ 0.30 } & \textbf{ 0.65 } & \textbf{ 1.07 } & \textbf{ 1.66 } & \textbf{ 2.31 } \\
    Our Model (25\%) & 0.53  & 0.94  & 1.38  & 2.27 & 3.14  \\
    Our Model (drop 3-frames) & \underline{0.35}  & 0.78  & \underline{1.23}  & \underline{1.77}  & \underline{2.42}  \\
    Our Model (drop 5-frames) & 0.46  & 0.92  & 1.29  & 1.86  & 2.74  \\
    Our Model (drop 8-frames) & 0.68  & 1.04  & 1.73 & 2.17  & 3.02  \\
    \bottomrule
    \end{tabular}%
    }
  \label{table_6}%
\end{table}%

\begin{table}[htbp]
  \centering
  \caption{Ablation results for various models on NGSIM and HighD.}
  \setlength{\tabcolsep}{4mm}
  \resizebox{0.95\linewidth}{!}{
    \begin{tabular}{c|ccccccc}
    \bottomrule
    \multicolumn{1}{c}{\multirow{2}[4]{*}{Dataset}} & \multirow{2}[4]{*}{Time (s)} & \multicolumn{6}{c}{ Model} \\
\cmidrule{3-8}    \multicolumn{1}{c}{} &       & A     & B     & C     & D     &E & F \\
    \hline
    \multirow{5}[2]{*}{NGSIM} & 1     & 0.47 & 0.45  & 0.41  & 0.43 &0.39 & \textbf{0.31} \\
          & 2     & 1.01  & 0.97  & 0.89  & 0.93 &0.88 & \textbf{0.81} \\
          & 3  &1.50   & 1.69  & 1.77  & 1.54  & 1.59 & \textbf{1.44} \\
          & 4     & 2.58  & 2.70  & 2.39  & 2.47  &2.27 & \textbf{2.08} \\
          & 5     & 3.32 & 3.41  & 3.10  & 3.26  &2.98 & \textbf{2.85} \\
    \hline
    \multirow{5}[2]{*}{HighD} & 1     & 0.06  & 0.0  & 0.04  & 0.05 &0.05& \textbf{0.04} \\
          & 2     & 0.13  & 0.14  & 0.10 & 0.12 &0.11 & \textbf{0.09} \\
          & 3     & 0.25  & 0.25  & 0.21  & 0.24 & 0.22 & \textbf{0.19} \\
          & 4     & 0.40  & 0.42  & 0.36  & 0.38 & 0.36 & \textbf{0.31} \\
          & 5     & 0.58  & 0.67  & 0.51  & 0.55  &0.49 & \textbf{0.45} \\
   \toprule
    \end{tabular}%
    }
  \label{table_5}%
\end{table}%

\begin{figure}[t]
  \centering  \includegraphics[width=0.95\linewidth]{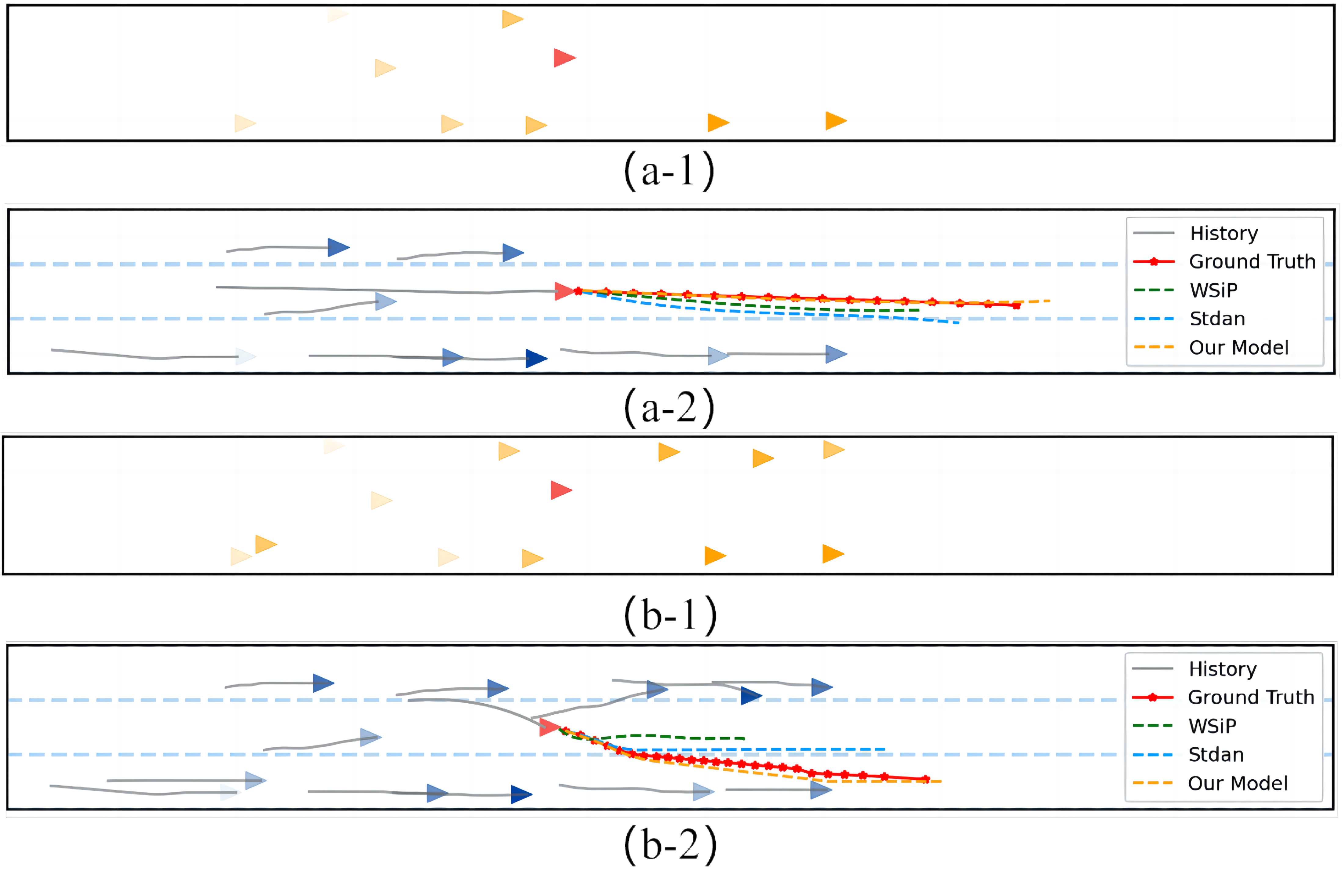} 
  \caption{Visual insights from the NGSIM dataset depict two complex driving scenarios: (a) driving straight and (b) merging to the right. The target vehicle is marked by red triangles, with surrounding vehicles represented by other triangles. Subfigures (a-1) and (b-1) present the outcomes of driver behavior profiling, whereas (a-2) and (b-2) illustrate the perceived safety metrics for each vehicle alongside their predicted trajectories. A vehicle shaded in deeper brown signals potential aggressive behavior, while a darker blue shade indicates a heightened risk to the target vehicle and vice versa.}
  \label{case} 
\end{figure}
\subsection{Qualitative Results}
It is an intrinsic understanding that vehicles in closer proximity often pose greater risks. However, an aspect frequently neglected by many models is that even distant vehicles, based on their driving behavior, can significantly influence the ego vehicle's trajectory. In our observations, driving behavior substantially impacts risk evaluations. Specifically, drivers with a more aggressive demeanor tend to escalate the risk levels. Drawing a direct comparison, our model's predictions stand out in their precision and alignment with reality, especially when juxtaposed against its counterparts like Stdan and WSiP. Taking the right merge scenario as a case in point: where Stdan and WSiP show discrepancies, it forecasts with minimal deviation, resonating closely with the ground truth. This differential is not just a testament to our model's accuracy but also its adeptness at capturing the nuanced effects of driving behaviors, a facet that seems muted in other models.

\subsection{Ablation Studies} \label{Ablation Studies}
We conducted a meticulous ablation study to evaluate the individual contributions of the components for our model. The summarized results are presented in Table \ref{table_5}. Specifically, Model F, which incorporates all components, consistently outperforms other variations across all evaluation metrics, clearly demonstrating the collective value of these components in achieving optimal performance.
Model A, excluding the DBP within the Perceived Safety-Aware module, suffers a significant degradation in performance, especially for short-term predictions, with degradations of at least 13.9\% and 24.0\% on the NGSIM and HighD datasets, respectively. This underscores the critical role of DBP in improving trajectory prediction accuracy. Moreover, Model B, a reduced version of Model F without the Perceived Safety-Aware module shows a significant reduction in RMSE, especially for long-term predictions, with improvements of at least 19.6\% and 26.2\% on the NGSIM and HighD datasets, respectively. This highlights the importance of considering perceived safety factors in trajectory prediction, especially for long-term prediction. Model C, which uses absolute coordinates instead of relative positions, displays non-negligible reductions in prediction metrics, highlighting the importance of spatial relationships in achieving accuracy.
In addition, Model D, which lacks the Interaction-Aware module, shows performance losses of at least 6.5\% and 20.0\% on the NGSIM and HighD datasets for short-term prediction, and at least 12.6\% and 18.2\% for long-term prediction, respectively. Finally, Model E, with reduced multimodal prediction in the decoder, shows at least 4.4\% and 8.2\% performance degradation on the NGSIM and HighD datasets, respectively. 

\section{Conclusion}\label{Conclusion}
In the evolving landscape of autonomous driving, trajectory prediction remains a complex challenge. This work presents a groundbreaking approach rooted in cognitive insights, emphasizing the crucial role of perceived safety in driving behaviors. Our Perceived Safety-Aware Module harmoniously merges Quantitative Safety Assessment and Driver Behavior Profiling, offering a detailed perspective on safety perceptions in driving. Rigorous evaluations on NGSIM, HighD, and MoCAD highlight our model's robustness and adaptability, even under data constraints and data missing. In conclusion, our findings suggest a promising trajectory for future research, bridging computational strength with human cognition to drive safety and efficiency in AVs.

\section*{Acknowledgements}
This research is supported by the Science and Technology Development Fund of Macau SAR (File no. 0021/2022/ITP, 0081/2022/A2, 001/2024/SKL), and University of Macau (SRG2023-00037-IOTSC).
\bibliographystyle{named}
\bibliography{ijcai24}

\end{document}